 \documentclass[final,1p,times]{elsarticle}

\usepackage{amssymb}

\usepackage{algorithmic} 

\usepackage{amsthm,amsmath,amssymb}

\def\R{\mathbb{R}}

\newtheorem{example}{Example}[section] 

\newtheorem{algorithm}{Algorithm}[section]

\usepackage{amsthm,amsmath,amssymb}

\def\R{\mathbb{R}}

\def\sign{\texttt{sign}}

\def\diag{\texttt{diag}}

\newcommand{\Eq}{{\bf E}}

\newcommand{\Vq}{{\bf V}}
\newcommand{\vq}{{\bf v}}
\newcommand{\Uq}{{\bf U}}
\newcommand{\uq}{{\bf u}}

\newcommand{\Xq}{{\bf X}}

\newcommand{\Zq}{{\bf Z}}

\newcommand{\Pq}{{\bf P}}

\newcommand{\Yq}{{\bf Y}}

\newcommand{\Tq}{{\bf T}}

\begin{document}

\begin{frontmatter}



\title{Advanced Variations of Two-Dimensional Principal Component Analysis for Face  Recognition
}


\author[1,2]{Meixiang Zhao}
\ead{zhaomeixiang2008@126.com}

\author[2]{Zhigang Jia
} 
\ead{zhgjia@jsnu.edu.cn}

\author[3]{Yunfeng Cai}
\ead{caiyunfeng@baidu.com}

\author[2]{Xiao Chen}
\ead{forward@126.com}

\author[1]{Dunwei Gong 
 }
\ead{dwgong@vip.163.com}

\address[1]{School of Information and Control Engineering, 
China University of Mining and Technology,
Xuzhou 221116, China}

\address[2]{School of Mathematics and Statistics  \&  Jiangsu Key Laboratory of Education Big Data Science and
Engineering, \\ Jiangsu Normal University,
Xuzhou 221116, China}

\address[3]{Cognitive Computing Lab, Baidu Research, Beijing  100193, China}

\begin{abstract}
 The  two-dimensional principal component analysis (2DPCA)  has become one of the most powerful  tools of artificial intelligent algorithms.  In this paper, we review  2DPCA and its variations, and  propose  a  general ridge regression model to  extract features  from both row and column directions. To enhance the generalization ability of extracted features, a novel relaxed 2DPCA  (R2DPCA)  is proposed  with a new  ridge regression model. R2DPCA  generates a weighting vector with utilizing the label information, and  maximizes a relaxed criterion with applying an optimal algorithm to get the essential features.  The  R2DPCA-based approaches for  face recognition and image reconstruction are also proposed and the selected  principle components are weighted to enhance the role of main components. Numerical experiments on well-known standard databases indicate that R2DPCA has high generalization ability and  can achieve a higher recognition rate  than the state-of-the-art methods, including in  the deep learning methods such as  CNNs, DBNs, and DNNs.
\end{abstract}

\begin{keyword}
2DPCA\sep Ridge regression model  \sep Feature extraction \sep Face recognition\sep Color image reconstruction



\end{keyword}

\end{frontmatter}


\section{Introduction}
\noindent
Two-Dimensional Principal Component Analysis (2DPCA) \cite{yzfy04}  and its variations (e.g., \cite{zhzh05,lpy10,whwj13,wangj16,gxcdgl19}) are playing an increasingly important  role in the recently proposed deep learning frameworks such as 2DPCANet \cite{yuwu17,lwk19}. It is always expected that 2DPCA can extract the spacial information and the best features of 2-D samples which can improve the performance of dimensional reduction.   From the view of numerical linear algebra,  the principle of 2DPCA   is to find a subspace (called eigenfaces or features) on which the projected samples have the largest  variance. The reconstruction from such projection or extraction of lower dimension is in fact the optimal low-rank approximation  of the original sample. When applying 2DPCA to face recognition, we  compute the eigenfaces or features based on the training set, and fairly use them to compress  the training and testing samples before classification.  An implicitly natural  assumption is that the projected  samples from the testing set still have large variance on the computed subspace. This exactly depends on  the generalization ability of 2DPCA.  In this paper,  we will review 2DPCA and variations, and  present a new relaxed 2DPCA (R2DPCA) with perfections in three aspects: abstracting the features of  matrix samples in both row and column directions, being innovatively armed with generalization ability,  and  weighting the main components  by corresponding eigenvalues.
Especially, R2DPCA  utilizes the label information of training data, and not only aims to enlarge the variance of projections of training samples.

The principal component analysis (PCA) \cite{Jolliffe04,TP91}, has become one of the most powerful  approaches  of face recognition \cite{siki87,kisi90,tupe91, zhya99,pent00}. Recently, many robust PCA (RPCA) algorithms are proposed with improving the quadratic formulation, which renders PCA vulnerable to noises, into  $L_1$-norm on the objection function, e.g., $L_1$-PCA \cite{keka05}, $R_1$-PCA \cite{dzhz06}, and PCA-$L_1$ \cite{kwak08}. Meanwhile, sparsity is also introduced into PCA algorithms, resulting in a series of sparse PCA (SPCA) algorithms \cite{zht06,agjl07,shhu08,wth09}. A newly proposed robust SPCA (RSPCA) \cite{mzx12} further applies $L_1$-norm both in objective and constraint functions of PCA, inheriting the merits of robustness and sparsity. Observing that $L_2$-, $L_1$-, and $L_0$-norms are all special $L_p$-norm,  it is natural to impose $L_p$-norm on the objection or/and constraint functions, straightforwardly; see  PCA-$L_p$ \cite{kwak14} and generalized PCA (GPCA) \cite{lxzzl13} for instance.

To preserve the spatial structure of face images, two dimensional PCA (2DPCA), proposed by Yang et al. \cite{yzfy04},   represents face images with two dimensional matrices rather than one dimensional vectors.  The computational problems bases on 2DPCA are of much smaller scale than those based on PCA, and the difficulties caused by rank defect are also  avoided in general. This image-as-matrix method offers insights for improving above  RSPCA,  PCA-$L_p$, GPCA,  etc.  As typical examples, the $L_1$-norm-based 2DPCA (2DPCA-$L_1$) \cite{lpy10} and 2DPCA-$L_1$ with sparsity (2DPCA$L_1$-S) \cite{whwj13} are improvements of PCA-$L_1$ and RSPCA, respectively, and the generalized 2DPCA (G2DPCA) \cite{wangj16} imposes $L_p$-norm on both objective and constraint functions of 2DPCA. 
Recently, the quaternion 2DPCA is proposed in \cite{jlz17} and applied to color face recognition, where the red, green and blue channels of a color image is encoded as three imaginary parts of a pure  quaternion matrix.  To arm the quaternion 2DPCA with the generalization ability,  Zhao, Jia and Gong \cite{zjg17} proposed the sample-relaxed quaternion 2DPCA with applying  the label information (if known) of training samples. The structure-preserving algorithms of quaternion eigenvalue decomposition and singular value decomposition can be found in \cite{jwl13,mjb18,jmz17,jwzc18, jns18b,jns18a}.  

Linear Discriminant Analysis (LDA) is another powerful feature extraction algorithm  in pattern recognition and computer vision. Since LDA often suffers from the small sample size (3S) problem,  some  effective approaches have been proposed, such as   PCA + LDA \cite{bhk97},  orthogonal LDA \cite{ye05}, LDA/GSVD \cite{hopa04}, and LDA/QR \cite{yeli05}. Because of  the advantages over the singularity problem and the computational cost, 2DLDA and its variants have recently attracted much attention from researchers  (e.g., \cite{lls08,kora05,xsa05,yyfz03,liyu05,cckkl06}). With applying the label information, the LDA-like methods are intend to compute the discriminant vectors which maximize the ratio of the between-class distance to the within-class distance.

PCA, 2DPCA and their variations are unsupervised methods,  without applying the potential or known  label information of samples.  Their features are calculated based on the training set and thus maximize the scatter of projected training samples.  The scatter of projected testing samples are not surely optimal, and certainly, so are the whole (training and testing) projected samples.  Inspired by this observation,   we present a new relaxation 2DPCA (R2DPCA).  This approach is a generalization of G2DPCA \cite{wangj16}, and will reduce to G2DPCA if the label information is unknown or unused.
Remark that the projection of R2DPCA does not aim to maximize  the variance of training samples as 2DPCA, but intends to avoid the overfitting and to enhance the generalization ability. R2DPCA sufficiently utilizes the labels (if known) of training samples, and can enhance  the total scatter of whole projected samples (see Example \ref{ex:toy} for the indication). Different to the idea of LDA,  R2DPCA aims to apply the label  information to generate a weighting vector and to construct a weighted covariance matrix in the newly proposed approach of face recognition. Thus R2DPCA never suffers from the small sample size (3S) problem.

 Our contributions are in three aspects.
(1) We  present a new ridge regression model for  2DPCA  and variations by $L_p$ norm. 
Such model is general and  abstracts features of face images from both row and column directions.  
 With this model, 2DPCA and variations are combined with additional regularization on the solution to fit various real-world applications, with the great flexibility.
(2) A  novel relaxed 2DPCA  (R2DPCA) is proposed with a new  ridge regression model.  R2DPCA has the stronger generalization ability than  2DPCA, 2DPCA-L1 2DPCAL1-S and G2DPCA.   To the best of our knowledge, we are the first to introduce the label information  into  the 2DPCA-based algorithms.  We also  weight the selected  principle components by corresponding eigenvalues to enhance the role of main components. 
(3) The  R2DPCA-based approaches are presented for  face recognition and image reconstruction, and their effectiveness is verified by applying them on practical face image databases.  They are indicated to perform better than the deep learning methods such as DNNs, DBNs and CNNs in the numerical examples.

The rest of this paper is organized as follows.
 In Section \ref{s:face}, we recall  2DPCA, 2DPCA-L1 2DPCAL1-S and G2DPCA, and present a ridge regression model to gather them together.  Their improved versions are also proposed. 
  In Section \ref{s:r2dpca}, we present a new relaxed two dimensional  principal component analysis (R2DPCA) and  the optimal algorithms.
We also present  the R2DPCA-based approaches for face recognition and image reconstruction.
  In Section \ref{s:exs}, we compare the R2DPCA with the state-to-the-art approaches, and indicate the efficiency of the R2DPCA . 
  In Section \ref{s:conclusion}, we sum up the contributions of this paper.

\section{The General Ridge Regression Model of  2DPCA and Variations}\label{s:face}
The two-dimensional principle component analysis (2DPCA)  has become one of the most popular and powerful methods in data sciences, especially in image recognition. Several  deep learning frameworks, which  rely hugely  on the wonderful properties of 2DPCA,  have achieved a high level performance in data analysis. This motivates us to develop a general model of 2DPCA and variations,  providing a  feasible way to embed them into artificial intelligence algorithms.  In this section, we firstly present a ridge regression model of the improved  2DPCA, and then analyze the relationship of the state-of-the-art variations of 2DPCA.

\subsection{A New Ridge Regression Model of Improved 2DPCA}\label{s:enhances2dpca}
The objective of  2DPCA is to find  left and/or right orthonormal  bases vectors so that the projected  matrix samples have the largest scatter after projection. Suppose  that  there are $n$ training matrix samples $ \textbf{X}_{1},\textbf{X}_{2},\ldots,\textbf{X}_{n} \in\mathbb{R}^{h\times w}$, where $h$ and $w$ denote the height and width of images, respectively.  Their mean value is  $ \Psi=\frac{1}{n}\sum\limits_{i=1}^{n}\Xq_i$. Let  $\Uq=[u_1,\cdots,u_{k_1}]\in\mathbb{R}^{h\times k_1}$ and $\Vq=[v_1,\cdots,v_{k_2}]\in\mathbb{R}^{w\times k_2}$ gather the left and right  optimal  basis vectors as columns, respectively.  Then  the $i$-th projected  matrix sample is defined by  $\Pq_i=\Uq^T(\Xq_i-\Psi)\Vq$.  The improved  2DPCA seeks optimal $\Uq$ and $\Vq$ that minimize the scatter of the  projected  matrix samples. This scatter is  characterized as 
 \begin{equation}\label{e:Juv}
 J(\Uq,\Vq)={\rm Tr}(\Uq^T \Eq_1\Uq) +  {\rm Tr}(\Vq^T \Eq_2\Vq), 
 \end{equation}
 where 
 \begin{equation}\label{e:E1E2}
\Eq_1=\frac{1}{n}\sum_{i=1}^{n}(\Xq_i-\Psi)(\Xq_i-\Psi)^T\in\mathbb{R}^{h\times h},\ \Eq_2=\frac{1}{n}\sum_{i=1}^{n}(\Xq_i-\Psi)^T(\Xq_i-\Psi)\in\mathbb{R}^{w\times w},
 \end{equation}
denote the  covariance matrices of input samples on  column and row directions, respectively. Both $\Eq_1$ and $\Eq_2$ are symmetric and semi-definite  matrices.  Since $\Uq$ and $\Vq$ are of full column rank, ${\rm Tr}(\Uq^T \Eq_1\Uq)$ and ${\rm Tr}(\Vq^T \Eq_2\Vq)$ are nonnegative. 
 Here, ${\rm Tr}(\cdot)$ represents the trace of a matrix.
 Thus, a new ridge regression model for  the improved
 2DPCA is  proposed as
 \begin{subequations}\label{e:argJ}
 \begin{align}\label{e:argJ1}
 (\widehat{\Uq},\widehat{\Vq})&={\rm\bf arg~max}~J(\Uq,\Vq)\\
\label{e:argJ2}{\rm s.t.}&~\Uq^T\Uq=I_{k_1}, ~\Vq^T\Vq=I_{k_2}.
 \end{align}
 \end{subequations} 
 To solve the optimal problem \eqref{e:argJ}, we need compute the eigenvalue problems of $\Eq_1$ and $\Eq_2$.  See Algorithm \ref{m:2DQPCA} for the detail.  
 \begin{center}
   \fbox{%
\parbox{0.85\textwidth}{
\begin{algorithm}[\bf Improved 2DPCA]\label{m:2DQPCA}
Input $n$  matrix samples, $\Xq_1,\cdots,\Xq_n\in\mathbb{R}^{h\times w}~(h,w\ge1)$, and two dimensions $k_1$ and $k_2$. Output the left and right  optimal  bases  $\Uq=[u_1,\cdots,u_{k_1}]\in\mathbb{R}^{h\times k_1}$ and $\Vq=[v_1,\cdots,v_{k_2}]\in\mathbb{R}^{w\times k_2}$.
\begin{itemize} 
\item[$(1)$] 
Compute the  {\it  covariance matrices} of training samples on column and row directions $\Eq_1$ and $\Eq_2$ as in \eqref{e:E1E2}.

\item[$(2)$] Compute the $k_1$  largest eigenvalues of $E_1$ and the corresponding   eigenvectors, denoted as $(\lambda_1, u_1),$ $\ldots,$ $(\lambda_{k_1}, u_{k_1})$.  Let  $\Uq=[u_1,\cdots,u_{k_1}]$.

\item[$(3)$] Compute the $k_2$  largest eigenvalues of $\Eq_2$ and the corresponding   eigenvectors, denoted as $(\lambda_1, v_1),$ $\ldots,$ $(\lambda_{k_2}, v_{k_2})$.  Let  $\Vq=[v_1,\cdots,v_{k_2}]$.
\end{itemize}
\end{algorithm}
}
}
\end{center}

\subsection{Improved Variations of 2DPCA and Optimal Algorithms}
Based on the idea in  Section \ref{s:enhances2dpca}, we present the improved versions of 2DPCA \cite{yzfy04},  2DPCA-$L_1$ \cite{lpy10}, 2DPCA$L_1$-S  \cite{whwj13}, and G2DPCA  \cite{wangj16}, whose ridge regression models are proposed in  the forms of computing the first projection vector.  

 Without loss of generality, we assume that the training samples are mean-centered, i.e., $\frac{1}{n}\sum^{n}\nolimits_{i=1}\textbf{X}_{i}=0$; otherwise, we will replace $\textbf{X}_i$ by $\textbf{X}_i-\frac{1}{n}\sum^{n}\nolimits_{i=1}\textbf{X}_i$.
After obtaining first $k$  left and right  projection vectors $\textbf{U}=[\textbf{u}_{1},\textbf{u}_{2},\ldots,\textbf{u}_{k}]$ and $\textbf{V}=[\textbf{v}_{1},\textbf{v}_{2},\ldots,\textbf{v}_{k}]$, the $(k+1)$-th left and right projection vectors $\textbf{u}_{k+1}$ and $\textbf{v}_{k+1}$ can be calculated similarly on deflated samples:
\begin{equation}\label{11}
  \textbf{X}_{i}^{R}=\textbf{X}_{i}(\textbf{I}-\textbf{V}\textbf{V}^{T}),\   \textbf{X}_{i}^{L}=\textbf{X}_{i}^T(\textbf{I}-\textbf{U}\textbf{U}^{T}),
\end{equation}
where  $i=1,2,\ldots,n$.
The ridge regression models of improved 2DPCA and variations find the first left and right projection vectors $\textbf{u}\in\mathbb{R}^{h}$ and $\textbf{v}\in\mathbb{R}^{w}$ by solving the optimization problem with equality constraints as follows.
\begin{itemize}
\item
The improved 2DPCA:
\begin{equation}\label{e:enhanced2DPCA}
  \max\limits_{\textbf{u}\in\mathbb{R}^{h},~\textbf{v}\in\mathbb{R}^{w}}\sum\limits^{n}_{i=1}\|\textbf{X}_{i}^L\textbf{u}\|_{2}^{2}+\|\textbf{X}_{i}^R\textbf{v}\|_{2}^{2}, \  s.t. ~\|\textbf{u}\|_{2}=1~{\rm and}~\|\textbf{v}\|_{2}=1.
\end{equation}
\item
 The improved 2DPCA-$L_1$:
\begin{equation}\label{e:improved2DPCAL1}
   \max\limits_{\textbf{u}\in\mathbb{R}^{h},~\textbf{v}\in\mathbb{R}^{w}}\sum\limits^{n}_{i=1}\|\textbf{X}_{i}^L\textbf{u}\|_{1}+\|\textbf{X}_{i}^R\textbf{v}\|_{1}, \  s.t. ~\|\textbf{u}\|_{1}=1~{\rm and}~\|\textbf{v}\|_{1}=1.
\end{equation}
\item
 The improved 2DPCA$L_1$-S:
\begin{equation}\label{e:improved2DPCAL1-S}
 \max\limits_{\textbf{u}\in\mathbb{R}^{h},~\textbf{v}\in\mathbb{R}^{w}}\sum\limits^{n}_{i=1}\|\textbf{X}_{i}^L\textbf{u}\|_{1}+\|\textbf{X}_{i}^R\textbf{v}\|_{1}, \  s.t. ~\|\textbf{u}\|_{2}=1,\|\textbf{v}\|_{2}=1, {\rm and}~\|\textbf{u}\|_{1},\|\textbf{v}\|_{1}\le c,
\end{equation}
where $c$ is a positive constant.
\item
The improved G2DPCA:
\begin{equation}\label{e:improvedG2DPCA}
  \max\limits_{\textbf{u}\in\mathbb{R}^{h},~\textbf{v}\in\mathbb{R}^{w}}\sum\limits^{n}_{i=1}\|\textbf{X}_{i}^L\textbf{u}\|_{s}^{s}+\|\textbf{X}_{i}^R\textbf{v}\|_{s}^{s}, \  s.t. ~\|\textbf{u}\|_{p}^p=1~{\rm and}~\|\textbf{v}\|_{p}^p=1,
\end{equation}
where $s\geq1$ and $p>0$.  
\end{itemize}

Since  two independent variables  in models \eqref{e:enhanced2DPCA}-\eqref{e:improvedG2DPCA}  are separated,  it is  appropriate to solve  $\uq$ and $\vq$ separately by optimal algorithms.   Taking the improved G2DPCA  \eqref{e:improvedG2DPCA} for instance,  the first projection vector $\textbf{w}$ ($=\uq$ or $\vq$) is computed by solving the optimization problem with equality constraints:
\begin{equation}\label{9}
  \max\limits_{\textbf{w}}\sum\limits^{n}_{i=1}\|\textbf{Y}_{i}\textbf{w}\|_{s}^{s}, \  s.t. \|\textbf{w}\|_{p}^{p}=1
\end{equation}
where $\Yq=\Xq_i^{L}$ or $\Xq_i^{R}$, $s\geq1$ and $p>0$.  Depending on the value $p$, the projection vector $\textbf{w}$ can be updated in two different ways.
 If $p\geq1$,
\begin{subequations}\label{e:pmethod}
\begin{align}
         &\textbf{w}^{k+1}=\sum\limits^{n}_{i=1}\textbf{Y}_{i}^{T}[|\textbf{Y}_{i}\textbf{w}^{k}|^{s-1}\circ \sign(\textbf{Y}_{i}\textbf{w}^{k})],\\
         &\textbf{w}^{k+1}=|\textbf{w}^{k+1}|^{q-1}\circ \sign(\textbf{w}^{k+1}),\ \textbf{w}^{k+1}=\frac{\textbf{w}^{k+1}}{\|\textbf{w}^{k+1}\|_{p}},
     \end{align}
 \end{subequations}
where $q$ satisfies $1/p+1/q=1$, $\circ$ denotes the Hadamard product, i.e., the element-wise product between two vectors.
If  $0<p<1$,
\begin{subequations}\label{e:pmethod1}
     \begin{align}
        &\textbf{w}^{k+1}=\sum\limits^{n}_{i=1}\textbf{Y}_{i}^{T}[|\textbf{Y}_{i}\textbf{w}^{k}|^{s-1}\circ \sign(\textbf{Y}_{i}\textbf{w}^{k})],\\
     &\textbf{w}^{k+1}=|\textbf{w}^{k}|^{2-p}\circ \textbf{w}^{k+1},\textbf{w}^{k+1}=\frac{\textbf{w}^{k+1}}{\|\textbf{w}^{k+1}\|_{p}}.
     \end{align}
 \end{subequations}
 
Notice that if the terms containing variable $\uq$ are omitted, the optimal models \eqref{e:enhanced2DPCA}-\eqref{e:improvedG2DPCA} are exactly the ridge regression models of well known 2DPCA \cite{yzfy04}, 2DPCA-$L_1$ \cite{lpy10}, 2DPCA$L_1$-S  \cite{whwj13}, and G2DPCA \cite{wangj16} algorithms.

\section{ Relaxed Two-Dimensional Principal Component Analysis}\label{s:r2dpca}
\noindent
2DPCA is an unsupervised methods and overlooks the potential or known  label information of samples.  The abstracted  features  maximize the scatter of projected training samples, and are implicitly expected to maximize (not surely)  the scatter of projected testing samples as well. In this section,  we present a new  relaxed two-dimensional principal component analysis (R2DPCA) method by $L_p$-norm  to avoid the overfitting and to enhance the generalization ability.  In large amount of experiments, R2DPCA sufficiently utilizes the labels (if known) of training samples, and can enhance  the total scatter of whole projected samples.  Interestingly, R2DPCA never suffers from the small sample size (3S) problem as supervised method such as LDA.  Now we introduce the  R2DPCA  from two parts:  weighting vector and objective function relaxation.

\subsection{Weighting vector}\label{ss:rv}

Suppose that training samples can be partitioned into $m$ classes and each class contains $n_j$  samples:
\begin{equation}\label{e:samplesmclasses}
\textbf{X}_1^1, \cdots, \textbf{X}_{n_1}^1\ | \ \textbf{X}_1^2, \cdots, \textbf{X}_{n_2}^2\ |\  \cdots \ |\  \textbf{X}_1^m, \cdots, \textbf{X}_{n_m}^m,
\end{equation}
where $\textbf{X}_i^j$ denotes the $i$-th sample of the $j$-th class, $i=1,\ldots, n_j$, $j=1,\ldots, m$.
Define the mean of training samples from the $j$-th class as
 $\Psi_j=\frac{1}{n_{j}}\sum \limits_{i=1}^{n_{j}}\textbf{X}_i^{j}\in\R^{h\times w},$
 and  
  the $j$-th within-class covariance matrix of the training set  as
$
\textbf{C}_j=\frac{1}{n_{j}}\sum\limits_{i=1}^{n_{j}}(\textbf{X}_i^j-\Psi_j)^T(\textbf{X}_i^j- \Psi_j)\in\R^{w\times w},
$
where $j=1,\ldots,m$, $\sum_{j=1}^{m}n_j=n$ and $i=1,\ldots,n_j$. The within-class covariance matrix $\textbf{C}_j$ is a symmetric and positive semi-definite  matrix. 
Its maximal eigenvalue, denoted by $\lambda_{\max}(\textbf{C}_j)$, represents the variance of training samples $\textbf{X}_1^j, \ldots, \textbf{X}_{n_j}^j$
 in the principal component. The larger $\lambda_{\rm max}(\textbf{C}_j)$ is,  the better scattered the training samples of $j$-th class are.  If $\lambda_{\max}(\textbf{C}_j)=0$ then all of training samples from the $j$-th class are same, and then the contribution of the $j$-th class  to the covariance matrix of training set should be reduced. To this aim, we define a {\it weighting vector} of training classes,
 \begin{equation}\label{e:v}
\omega=[\omega_1,\cdots,\omega_m]^T\in\mathbb{R}^m, \ 
   \omega_j=\frac{f(\lambda_{\rm max}(\textbf{C}_j)) }{\sum_{i=1}^{m}f(\lambda_{\rm max}(\textbf{C}_i)) },
 \end{equation}
 where $ \omega_j$  is a weighting factor of the $j$-th class with a function, $f:~\R\rightarrow \R^{+}$.
The computation of the weighting vector  is proposed in Algorithm \ref{a:rv}.
 \begin{center}
   \fbox{%
\parbox{0.85\textwidth}{
\begin{algorithm}[\bf Weighting Vector]\label{a:rv}
 Input $n$ training samples as in \eqref{e:samplesmclasses},  the number of classes  $m$, the number of samples in each class contains $n_j$ and the dimension $w$. Output the  weighting vector $\omega$.

\begin{algorithmic}[section]\small
\STATE {\bf function} $\omega=\texttt{weightvec}(\textbf{X}_1, \textbf{X}_2,\cdots, \textbf{X}_n,m,w,n_j)$
\FOR {$j=1,2,\cdots,m$}
\STATE $\textbf{C}_j=\texttt{zeros}(w,w)$
\STATE $\textbf{M}_j=\frac{1}{n_j}(\textbf{X}_1^j+\cdots+\textbf{X}_{n_j}^j)$
\FOR{$i=1,2,\cdots,n_j$}
\STATE $\textbf{C}_j=\textbf{C}_j+{(\textbf{X}_i^j-\Psi_j)}'*(\textbf{X}_i^j-\Psi_j)$
\ENDFOR
\STATE $\textbf{C}_j=\textbf{C}_j/n_j;$
\STATE {\bf Compute relaxation vector $\omega$ defined as in} \eqref{e:v}
\ENDFOR
\end{algorithmic}
\end{algorithm}
}
}
\end{center}

\subsection{Objective function relaxation}\label{ss:ofr}
With the computed weighting vector $\omega$ in hand,  we define a {\it relaxed criterion} as
\begin{equation}\label{e:gtsc}
J(\textbf{u},\textbf{v})=\gamma\textbf{G}(\textbf{u},\textbf{v})+(1-\gamma)\widetilde{\textbf{G}}(\textbf{u},\textbf{v}),
\end{equation}
where  $ \gamma\in[0,1]$ is a relaxation parameter,   $\textbf{u}\in\R^h$ and $\textbf{v}\in\R^w$ are unit vectors under $L_p$ norm, 
\begin{subequations}\label{e:GwideG}
\begin{align}\label{e:Guv}\textbf{G}(\textbf{u},\textbf{v})&:=\sum\limits^{n}_{i=1}\|\textbf{u}^T(\textbf{X}_{i}-\Psi)\|_{s}^{s}+\|(\textbf{X}_{i}-\Psi)\textbf{v}\|_{s}^{s},\\
\label{e:wideGuv}\widetilde{\textbf{G}}(\textbf{u},\textbf{v})&:=\sum\limits^{m}_{j=1}\sum\limits^{n_j}_{i=1}\left\|\frac{\omega_j}{n_j}\textbf{u}^T(\textbf{X}_i^j-\Psi)\right\|_{s}^{s}+\left\|\frac{\omega_j}{n_j}(\textbf{X}_i^j-\Psi)\textbf{v}\right\|_{s}^{s}.
\end{align}
\end{subequations}
The R2DPCA finds its first projection vectors $\textbf{u}\in\mathbb{R}^{h}$ and $\textbf{v}\in\mathbb{R}^{w}$ by solving the optimization problem with equality  constraints:
\begin{equation}\label{e:r2dcri}
\max\limits_{\textbf{u}\in\mathbb{R}^{h},\textbf{w}\in\R^w} 
J(\textbf{u},\textbf{v}),~~
 s.t. ~~\|\textbf{u}\|_p^p=1~\text{and}~\|\textbf{v}\|_p^p=1,
\end{equation}
where the criterion $J(\textbf{u},\textbf{v})$ is defined as in \eqref{e:gtsc}.
 Notice that the  relaxed criterion \eqref{e:r2dcri} reduces  to  \eqref{e:improvedG2DPCA} if $\gamma=1$, and thus,  the first projection vectors of R2DPCA and  G2DPCA are the same.
If first $k$ projection vectors $\textbf{U}=[\textbf{u}_{1},\textbf{u}_{2},\ldots,\textbf{u}_{k}]$ and $\textbf{V}=[\textbf{v}_{1},\textbf{v}_{2},\ldots,\textbf{v}_{k}]$ have been obtained, the $(k+1)$-th projection vectors $\textbf{u}_{k+1}$ and $\textbf{v}_{k+1}$ can be calculated similarly on the deflated samples, defined as in \eqref{11}.  At each iterative step,
 we also obtain the  maximal value of the objective function,
\begin{align*}
   f_k&=\gamma\sum\limits^{n}_{i=1}\|\textbf{u}_k^T(\textbf{X}_{i}-\Psi)^{R}\|_{s}^{s}+\|(\textbf{X}_{i}-\Psi)^{R}\textbf{v}_k\|_{s}^{s}
   \\
   &+(1-\gamma)\sum\limits^{m}_{j=1}\sum\limits^{n_j}_{i=1}\left\|\frac{\omega_j}{n_j}\textbf{u}_k^T(\textbf{X}_i^j-\Psi)^{R}\right\|_{s}^{s}+\left\|\frac{\omega_j}{n_j}(\textbf{X}_i^j-\Psi)^{R}\textbf{v}_k\right\|_{s}^{s}.
\end{align*}
Exactly,  the first $k$ optimal projection vectors of R2DPCA solve the optimal problem with equality constraints:
 \begin{equation}\label{e:r2d4ws}
\begin{array}{l}
\{\textbf{u}_1,\ldots,\textbf{u}_k,\textbf{v}_1,\ldots,\textbf{v}_k\}=\arg \max J(\textbf{u},\textbf{v})\\ [5pt]
\ {\rm s.t.}\ \
\left\{
\begin{array}{l}  \|\textbf{u}_i\|_p^p=1,~\|\textbf{v}_i\|_p^p=1,\\
 \textbf{u}_i^T\textbf{u}_j=0,~\textbf{v}_i^T\textbf{v}_j=0 ~i\neq j,
 \end{array}
~i,j=1,\cdots,k.
\right.
\end{array}
\end{equation}
Algorithm \ref{a:r2dpca} is presented to compute first  $k$ optimal left and right projection vectors.

Now we present the relationships among the improved 2DPCA, 2DPCA$L_1$, 2DPCA$L_1$-S , G2DPCA,  and R2DPCA.
It is obvious that 2DPCA and 2DPCA-$L_1$ are two special cases of G2DPCA.
2DPCA$L_1$-S originates from G2DPCA with $s=1$ and $p=1$ which leads to projection vector with only one nonzero element. Then the $L_2$-norm constraint is employed to fix this problem, resulting in 2DPCA$L_1$-S. On the other hand, G2DPCA with $s=1$ and $1<p<2$ behaves like 2DPCA$L_1$-S,  since the $L_p$-norm constraint in G2DPCA behaves like the mixed-norm constraint in 2DPCA$L_1$-S. R2DPCA is a generalization of G2DPCA.

 \begin{center}
   \fbox{%
\parbox{0.90\textwidth}{
\begin{algorithm}[\bf  Relaxed Two-Dimensional Principal Component Analysis (R2DPCA)]
\label{a:r2dpca}  %
\begin{algorithmic}[section]\small %
Input training samples  as in \eqref{e:samplesmclasses}
and parameters $s\in [1,\infty)$, $p\in(0,\infty]. $
\STATE (1) Computing the weighting vector, $\omega$, by Algorithm \ref{a:rv}.
\STATE (2) Compute the  covariance matrix  and the relaxed covariance matrix as in \eqref{e:GwideG}.
\STATE (3) Compute the left features 
 $\textbf{U}=[ \textbf{u}_1,\ldots,\textbf{u}_k]$, the right features  $\textbf{V}=[ \textbf{v}_1,\ldots,\textbf{v}_k]$,   and the variances  $\textbf{D}=\diag( f_1,\ldots,f_k)$, according to the relaxed criterion \eqref{e:gtsc}.
\FOR{$t=1,2,\cdots,r$}
\STATE Initialize $k=0$, $\delta=1$, arbitrary $\textbf{v}^0$ with $\parallel\textbf{v}^0\parallel_p=1$.
\STATE  $f_0=J(\textbf{u}^0,\textbf{v}^0)$ according to  \eqref{e:gtsc}.
\WHILE{$\delta> tol$}
\STATE  $\textbf{v}^{k+1}$ is computed in the following four cases.  $\textbf{u}^{k+1}$ is computed by the similar way. 
\STATE \quad $\textbf{w}^{k}=\gamma\sum\limits^{n}_{i=1}\textbf{X}_{i}^{T}[|\textbf{X}_{i}\textbf{v}^k|^{s-1}\circ \sign(\textbf{X}_i\textbf{v}^k)]+$
\STATE \quad\quad \quad \quad \quad $(1-\gamma)\sum\limits^{m}_{j=1}\sum\limits^{n_j}_{j=1}(\frac{v_j}{n_j}\textbf{X}_{i}^j)^{T}[|\frac{\omega_j}{n_j}\textbf{X}_{i}^j\textbf{v}^k|^{s-1}\circ \sign(\frac{\omega_j}{n_j}\textbf{X}_{i}^j\textbf{v}^k)]$.
\STATE \quad \textbf{Case 1:} $0<p<1$.
 \quad $\textbf{w}^k=|\textbf{v}^k|^{2-p}\circ\textbf{w}^k,$ \quad $\textbf{v}^{k+1}=\frac{\textbf{w}^k}{\parallel\textbf{w}^k\parallel_p}.$
\STATE \quad\textbf{Case 2:} $p=1$.
 \quad $j=\arg \max\nolimits_{i\in [1,w]}|w_{i}^{k}|,$
 \quad $ v_{i}^{k+1}=\left\{
\begin{aligned}
\sign(v_{j}^{k}), \ i=j,\\
0, \ i\neq j.\\
\end{aligned}
\right.$
\STATE \quad \textbf{Case 3:} $1<p<\infty$.
 \quad$q=p/(p-1),$
 \quad$\textbf{w}^k=|\textbf{w}^k|^{q-1}\circ \sign(\textbf{w}^k),$
 \quad$\textbf{v}^{k+1}=\frac{\textbf{w}^k}{\parallel\textbf{w}^k\parallel_p}.$
\STATE \quad\textbf{Case 4:} $p=\infty$.
 \quad$\textbf{v}^{k+1}=\sign(\textbf{w}^k).$
\STATE  $f_{k+1}=J(\textbf{u}^{k+1},\textbf{v}^{k+1})$ according to  \eqref{e:gtsc}.
\STATE $\delta=|f_{k+1}-f_{k}|/|f_k|.$
\STATE $k\leftarrow k+1.$
\ENDWHILE
\STATE 
$\textbf{U}\leftarrow [\textbf{U},\textbf{v}^{k}].$ 
 $\textbf{V}\leftarrow [\textbf{V},\textbf{v}^{k}].$
 $\textbf{D}=\diag(\textbf{D},f_k).$
\STATE   $\textbf{X}_i^R=\textbf{X}_i^R(\textbf{I}-\textbf{V}\textbf{V}^T), i=1,2,\cdots,n.$
\ENDFOR
\end{algorithmic}
\end{algorithm}
}
}
\end{center}

\subsection{Face recognition}\label{ss:eigenfaces}
Suppose we have computed the optimal projections,  $\widehat{\Uq}$ and $\widehat{\Vq}$, and the diagonal matrix $D$ by   R2DPCA. The R2DPCA approach for color face recognition is proposed in Algorithm \ref{m:sr2dpca}. 
\begin{center}
   \fbox{%
\parbox{0.85\textwidth}{
\begin{algorithm}[\bf R2DPCA approach for face recognition]\label{m:sr2dpca} 
Input the training set, $\{\Xq_1,\cdots,\Xq_n\}$, the optimal projections,  $\widehat{\Uq}$ and $\widehat{\Vq}$, and the set of face images to be recognized, $\mathbb{T}=\{\Tq_1,\cdots,\Tq_k\}$. Output the identity  vector of  $\mathbb{T}$, $r\in\mathbb{R}^k$. 
\begin{itemize} 
\item[$(1)$] Compute the features of $n$ training face images under $\widehat{\Uq}$ and $\widehat{\Vq}$ as 
\begin{equation*}\label{e:ps4fs}\widehat{\Xq}_i=\widehat{\Uq}^*(\Xq_i-\Psi)\widehat{\Vq}\in\mathbb{H}^{k_1\times k_2},\ i=1,\cdots,n.
\end{equation*}
\item[$(2)$]  Compute the  feature of each face images in $\mathbb{T}$, $\widehat{\Tq}_j=\widehat{\Uq}^*(\Tq_j-\Psi)\widehat{\Vq}$. 
\item[$(3)$] Solve the optimal problems 
$$r(j)={\bf arg~min}_{1\le i\le n} \|(\widehat{\Tq}_j-\widehat{\Xq}_i)D\|$$
 for $j=1,\cdots,k$.

\end{itemize}
 \end{algorithm}
 }
 }
 \end{center}
%

\subsection{Image reconstruction}\label{ss:imreconstruction}

The original digit image, $\Xq_i$, can be optimally  approximated by a low-rank reconstruction  from its feature,  $\widehat{\Xq}_i$.  Suppose that $\widehat{\Uq}^\bot \in\mathbb{R}^{h\times (h-k_1)}$ and $\widehat{\Vq}^\bot \in\mathbb{R}^{n\times (w-k_2)}$ are  the unitary complement of~$\widehat{\Uq}$ and $\widehat{\Vq}$.  For $i=1,\cdots,n$, the reconstructions are defined as 
 \begin{equation}\label{e:ps_ps2}\widehat{\Uq}\widehat{\Xq}_i\widehat{\Vq}^T=\Xq_i+\Pq, \end{equation}
 with $\Pq\in\{\widehat{\Uq}^\bot (\widehat{\Uq}^\bot)^T\Zq +\Zq\widehat{\Vq}^\bot (\widehat{\Vq}^\bot)^*|\Zq\in\mathbb{R}^{h\times w}\}$.  Here, the mean value of samples  is assumed to be zero for simplicity. 
 The image reconstruction rate of $\widehat{\Uq}\widehat{\Xq}_i\widehat{\Vq}^T$  is defined as follows
\begin{equation}\label{e:ratio}Ratio_s=1-\frac{\|\widehat{\Uq}\widehat{\Xq}_i\widehat{\Vq}^T-\Xq_i\|_2}{\|\Xq_i\|_2}=\frac{\|\Xq_i\|_2-\|\Pq\|_2}{\|\Xq_i\|_2}.\end{equation}
Note that $\widehat{\Uq}\widehat{\Xq}_i\widehat{\Vq}^T$ is always a good approximation of  $\Xq_i$. If  $k_1=h$ and $k_2=w$,   $\widehat{\Uq}$ and $\widehat{\Vq}$ are a unitary matrices and hence $Ratio_i=1$, which means $\widehat{\Uq}\widehat{\Xq}_i\widehat{\Vq}^T=\Xq_i$.

\section{Experiments }\label{s:exs}
\noindent
In this section, we present numerical experiments  to compare all advanced variations of 2DPCA, including in   the relaxed two-dimensional principle component analysis (R2DPCA),  with the state-of-the-art algorithms.  
%
%
The numerical experiments are performed with MATLAB-R2016 on a personal computer with Intel(R) Xeon(R) CPU E5-2630 v3 @  2.4GHz (dual processor) and RAM 32GB.

%

\begin{example}\label{example1}
In this experiment, we compare R2DPCA with  2DPCA, 2DPCA-$L_1$, 2DPCA$L_1$-S, and  G2DPCA on face recognition by utilizing three famous databases  as follows:
\begin{itemize}
\item Faces95 database\footnote{Collection of Facial Images:  Faces95.  http://cswww.essex. ac.uk/mv/allf aces/faces95.html} (1440 images from 72 subjects, twenty images per subject),
\item Color FERET database\footnote{The color Face Recognition Technology (FERET) database: https://www.nist.gov/itl/iad/image-group/color-feret-database.} (3025 images from 275 subjects, eleven images per subject),
\item Grey FERET database\footnote{Here we use the widely used cropped version of the FERET database. The size of each face image is $80\times 80$. 
  }  (1400 images from 200 subjects, seven images per subject).
\end{itemize}
All of face images are cropped and resized such that each image is  of 80$\times$80 size.
The basic setting is that  $10$ and $5$ face images  of each person  from  Faces95 and (color or grey) FERET face databases are randomly selected out as training samples, and the remaining ones are left for testing. 

We test the effect of numbers of chosen features on  the recognition accuracy. 
Let $s$ and $p$  be fixed as the optimal parameters in Table \ref{faces95table}.  The recognition accuracies of 2DPCA, 2DPCA-L1, 2DPCAL1-S, G2DPCA and R2DPCA with different feature numbers in the range of $[1,20]$ on the grey Feret databases are shown in Figure \ref{pca_like_greyFERET}. 
The recognition accuracies of G2DPCA and R2DPCA  on the Faces95  and Color Feret databases are shown in Figure \ref{faces95_experiment2}.

From the numerical results in Table \ref{faces95table} and Figures \ref{pca_like_greyFERET} and \ref{faces95_experiment2},  we can conclude that the classification accuracies of R2DPCA are higher and more stable than 2DPCA, 2DPCA-L1, 2DPCAL1-S and G2DPCA when the number of chosen features is large.   The recognition accuracies of G2DPCA and R2DPCA are the same when $k=1$, in which case neither relaxation nor weighting is necessary in G2DPCA.

\begin{table}[!t]
\caption{Recognition accuracies of five algorithms}\label{faces95table}
\center
\begin{tabular}{c|cc|cc}
 \hline
Algorithms
      & \multicolumn{2}{c|}{Face95} 
      &  \multicolumn{2}{c}{color Feret}       \\ 
& $Optimal parameters$ & $Accuracy$&$Optimal parameters$ & $Accuracy$\\ \hline
2DPCA&   $-$ & $0.8729$                              &   $-$ & $0.5982$ \\
2DPCA-$L_1$&   $-$ &$0.8708$                    &   $-$ &$0.5985$      \\
2DPCA$L_1$-S&  $\rho=-0.5$ &$0.8785$     &  $\rho=-0.3$ &$0.6236$  \\
G2DPCA&   $s=2.7,p=2.2$ &$0.9451$          &   $s=2.8,p=2.6$ &$0.6918$   \\
R2DPCA $(\gamma=0)$&   $s=1,p=2.2$ &$\textbf{0.9493}$  &   $s=3,p=2.2$ &$\textbf{0.7085}$  \\\hline
\end{tabular}
\end{table}

\begin{figure}[!h]
  \begin{center}
\includegraphics[height=0.45\textwidth,width=0.95\textwidth]{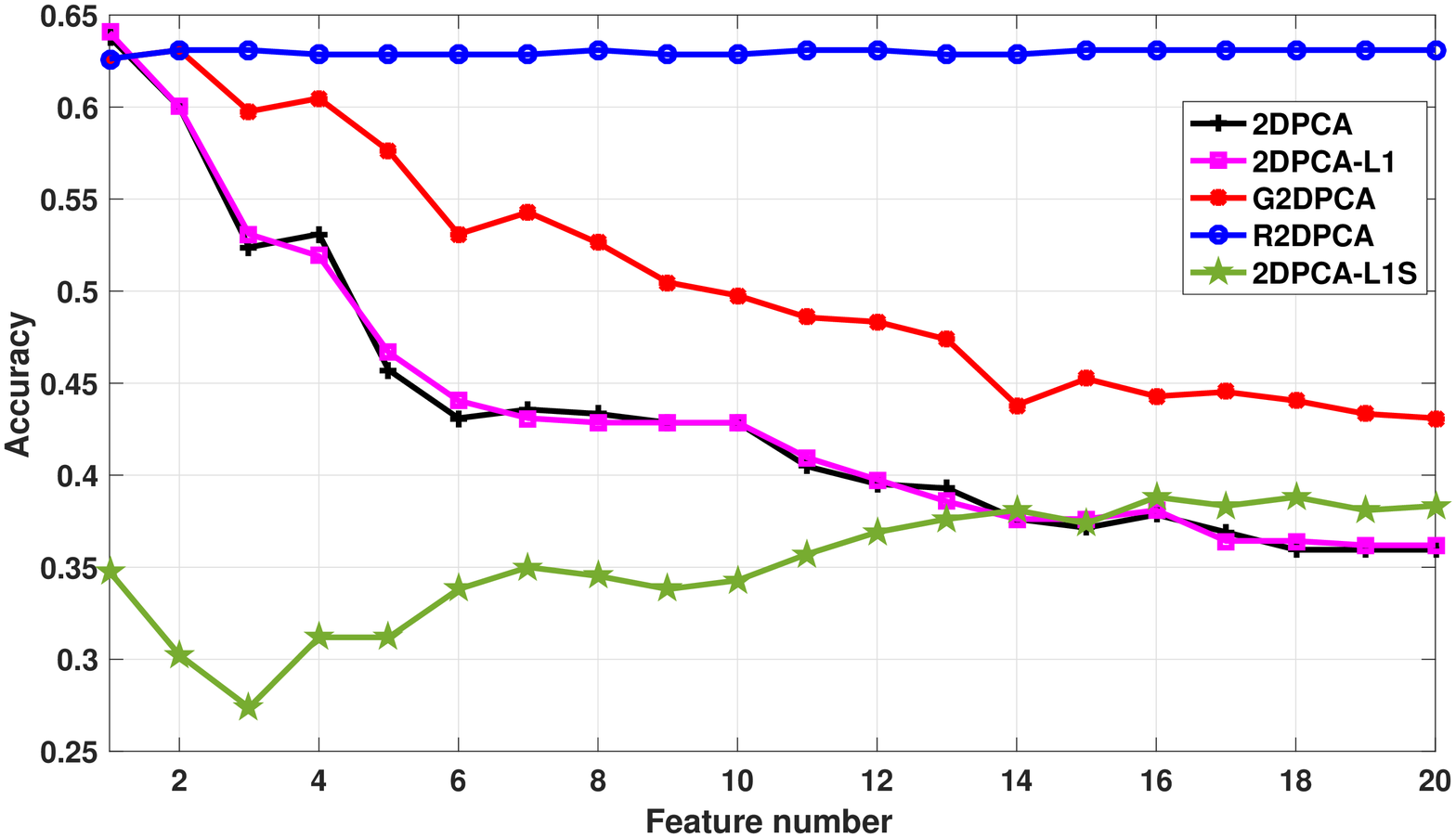}
  \end{center}
 \vskip-15pt \caption{\small Recognition accuracies of  2DPCA, 2DPCA-L1, 2DPCAL1-S, G2DPCA and R2DPCA with $k=[1:20]$ on the grey FETET database.}
\label{pca_like_greyFERET}
\end{figure}

\begin{figure}[!h]
  \begin{center}
\includegraphics[height=0.4\textwidth,width=0.9\textwidth]{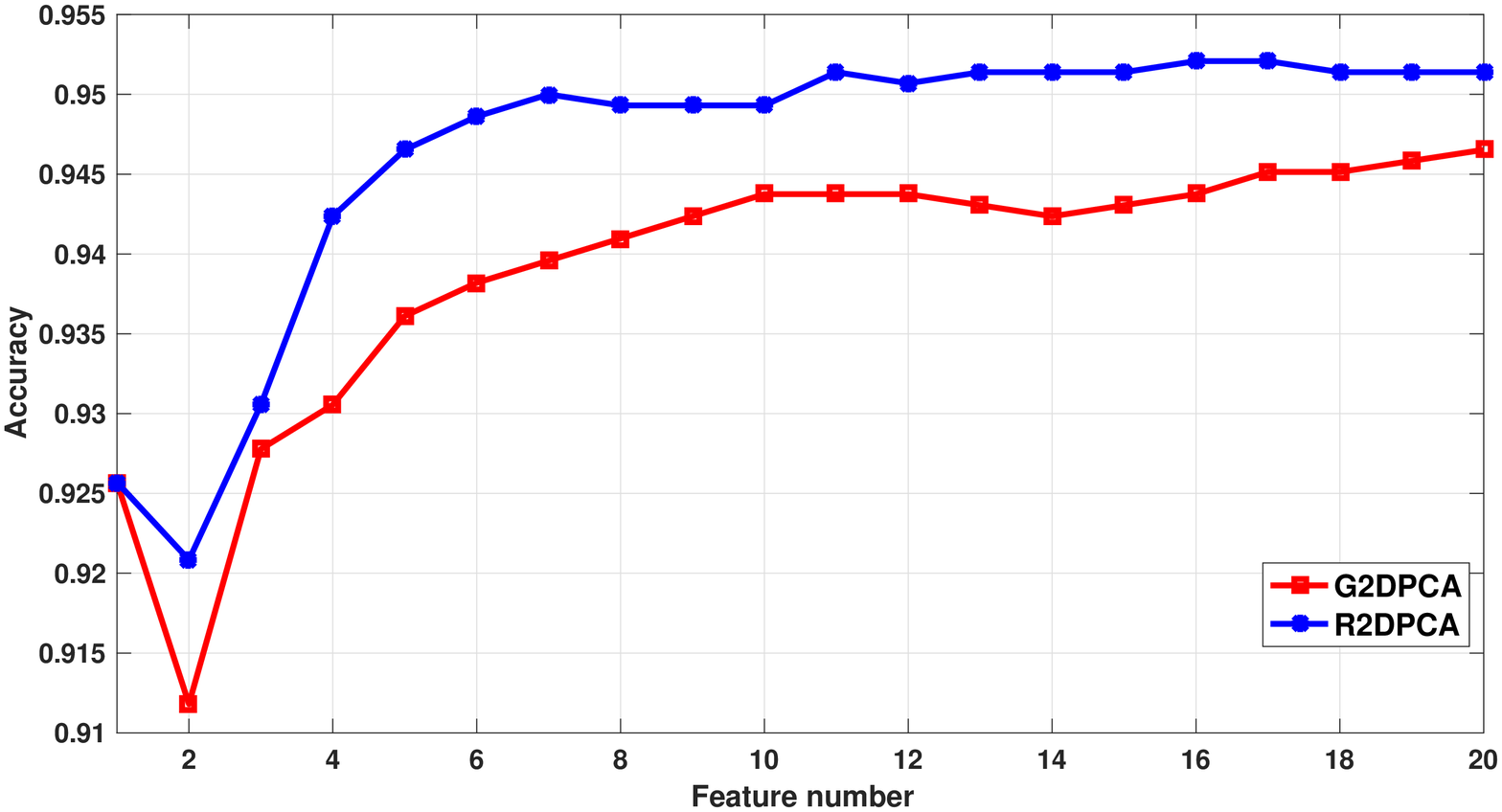}
\includegraphics[height=0.4\textwidth,width=0.9\textwidth]{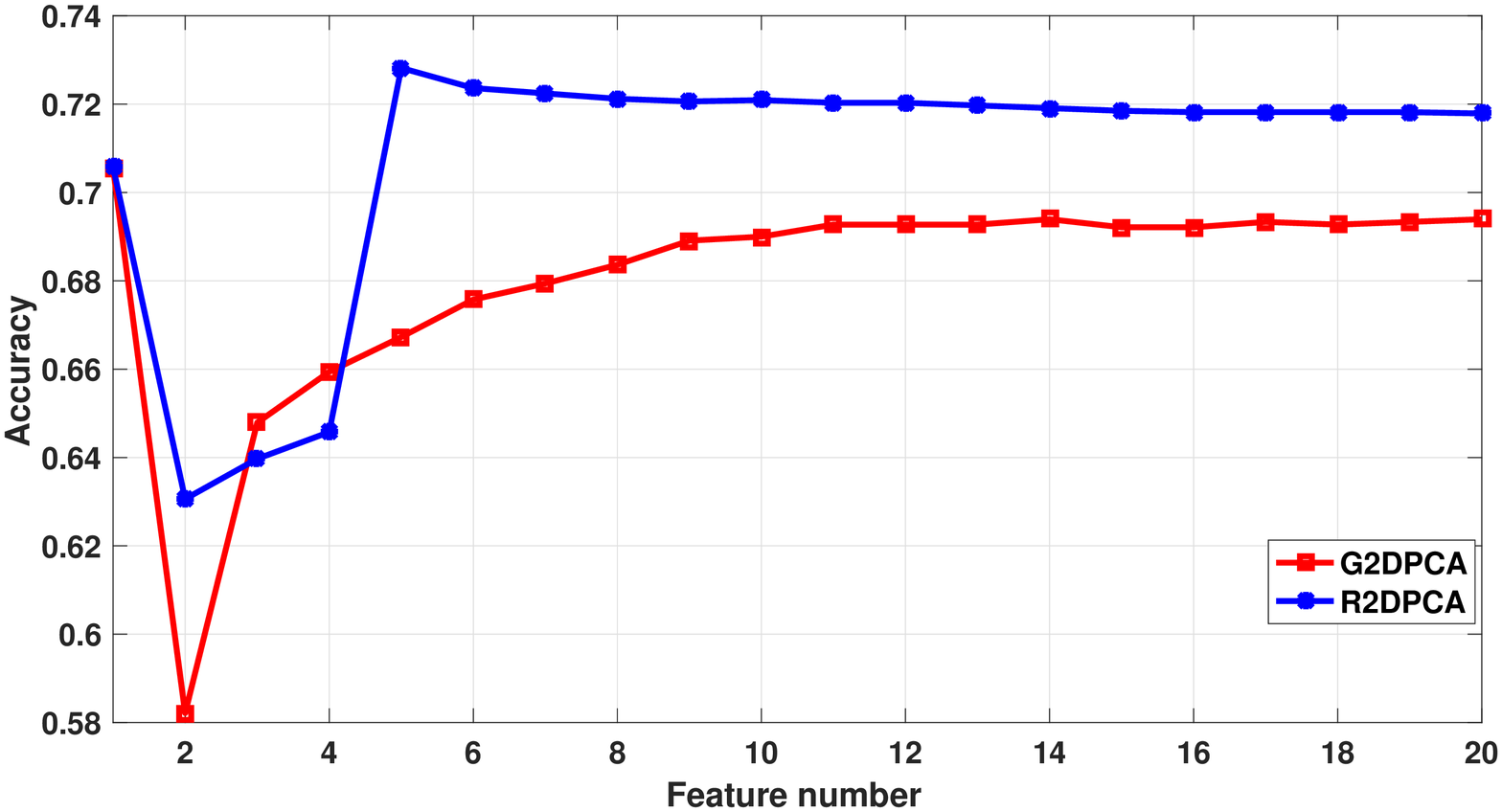}
  \end{center}
  \vskip-15pt \caption{Recognition accuracies of R2DPCA and G2DPCA with $k=[1:20]$ on the Faces95 (top) and Color FERET (below) databases.}
\label{faces95_experiment2}
\end{figure}

\end{example}

\begin{example}
In this experiment,  we compare R2DPCA with three most prominent deep learning primitives:  Convolutional Neural Networks (CNNs), Deep Belief Networks (DBNs) and Deep Neural Networks (DNNs). 
These methods are applied on the partial MNIST database of handwritten digits, which has a training set of $10000$ samples, and a test set of $1000$ samples. The size of each image is $28\times 28$ pixels.  
The  codes of CNNs, DBNs and DNNs are according to  \cite{palm12} and \cite{nielsen13}, and the settings are as follows.   

Deep Neural Networks (DNNs) are implemented by stacking layers of neural networks along the depth and width of smaller architectures. A four-layer neural network is used in our tests. The input layer of the network  contains $784=28\times 28$ neurons and the output layer contains $10$ neurons. The number of neurons in first and second hidden layers are set  by $n$ and $\ell$, where $n$ increases from $1$ to $12$ and  $\ell=14$ is fixed. All weights and biases are initialized randomly between $[0,1]$ and will be updated by error back propagation algorithm.  
The iteration will stop once the convergence condition is achieved. In our test, this condition is that if current accuracy of test samples is lower than last iteration more than three times.
 
Deep Belief Networks (DBNs) consist of a number of layers of Restricted Boltzmann Machines (RBMs) which are trained in a greedy layer wise fashion. The lower layer is same as the input layer in DNNs, and the top layer as the hidden layer. In our experiment, a four layers consisted of two RBMs are constructed. We set the number of hidden neurons in first RBM from $10$ to $120$, step $10$ and a fixed number of hidden neurons of $100$ in second RBM. Each RBM is trained in a layer-wise greedy manner with contrastive divergence. All weights and biases are initialized to be zero. Each RBM is trained on the full $10000$ images training set, using mini-batches of size $100$, with a fixed learning rate of $1$ for one epoch. One epoch is one full sweep of the data. Having trained the first RBM the entire training dataset is transformed through the first RBM resulting in a new $10000$ by $k$ $(k=10:10:120)$ dataset which the second RBM is trained on. Then the trained weights and biases are used to initialize a feed-forward neural net with $3$ layers of sizes $k-1000-10$, the last 10 neurons being the output label units. The feed-forward neural net is trained with sigmoid activation function using backpropagation. Here we set the mini-batches of size $100$ for one epoch using a fixed learning rate of $1$. At last the $1000$ test samples are performed in the feed-forward network and the maximum output unit are their labels.

Convolutional Neural Networks (CNNs) are feed-forward, back-propagate neural networks with a special architecture inspired from the visual system, consisting of alternating layers of convolution layers and sub-sampling layers. What is different is that CNNs work on the two dimensional data directly. In our experiment, we set two convolution layers and two sub-sampling layers. The first layer has k feature maps, where we set $k$ from $1$ to $12$, step $1$, connected to the single input layer through $k$ $5\times 5$ kernels. The second layer is a $2\times 2$ mean-pooling layer. The third layer has $12$ feature maps which are all connected to all k mean-pooling layers below through $12k$ $5\times 5$ kernels. The fourth layer is still a $2\times 2$ mean-pooling layer. After above steps the feature maps is concatenated into a feature vector which feeds into the final layer which consists of $10$ output neurons, corresponding to the $10$ class labels. 
The CNNs are trained with stochastic gradient descent on the training set, using mini-batches of size $50$, with a fixed learning rate of $1$ for one epoch. 
Putting test samples in the trained networks and comparing output with their true labels in order are  to get the recognition rate.

\begin{figure}[!h]
  \begin{center}
\includegraphics[height=0.45\textwidth,width=0.9\textwidth]{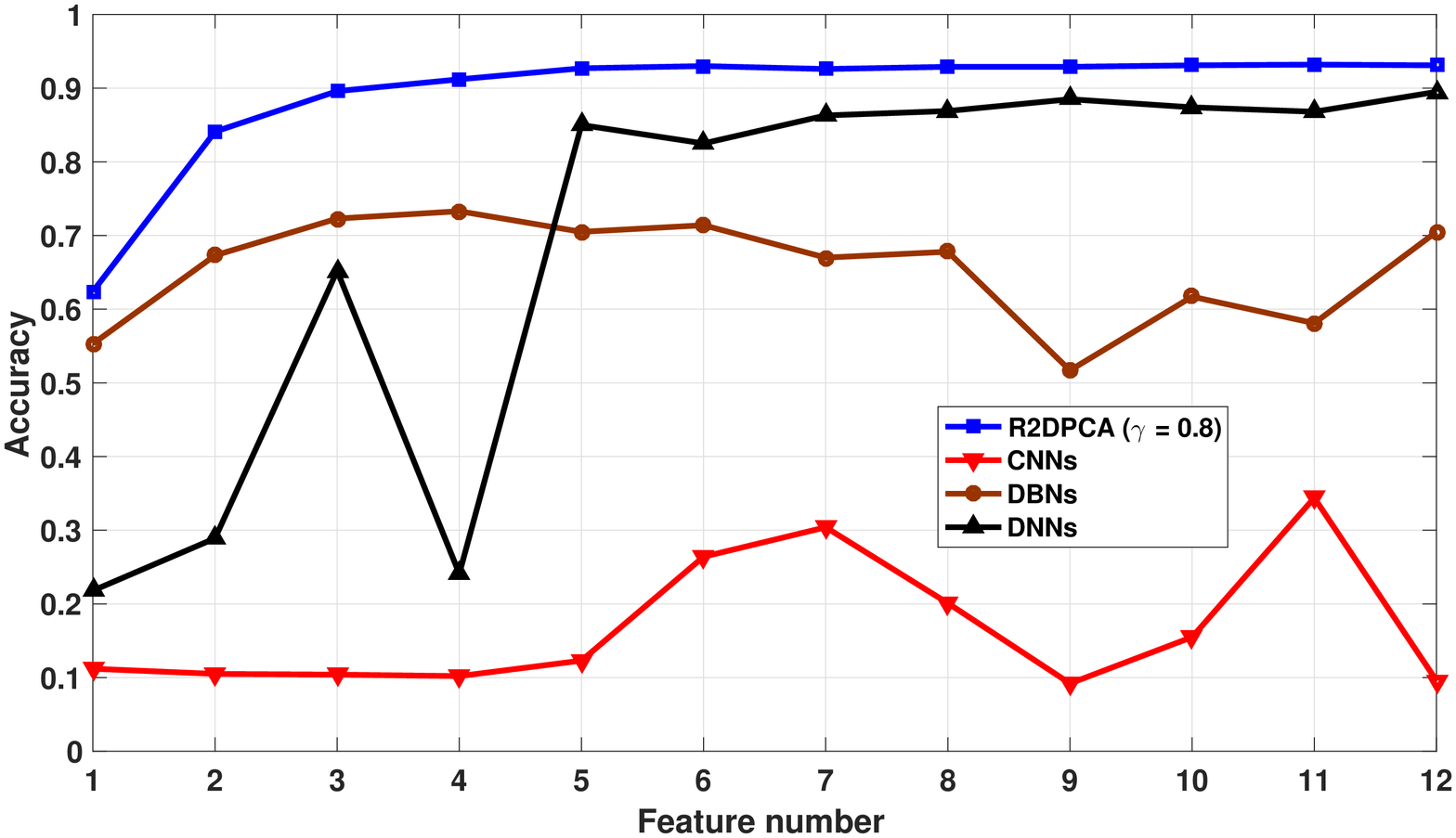}
  \end{center}
  \vskip-15pt \caption{Recognition accuracies on the MINIST database.}
\label{r2dpcaCNN}
\end{figure}

The numerical results are shown in Figure \ref{r2dpcaCNN}. We can see that  R2DPCA  has the better performance over  CNNs, DBNs and DNNs in the recognition accuracies.    It should be noticed that the recognition rates of CNNs, DBNs and DNNs can not achieve at high levels with small samples, but  will increase when the amount of training samples become larger. 
\end{example}

\begin{example}\label{ex:toy} In this experiment, we indicate the generalization ability of R2DPCA.   Let $4n$ randomly generated points  be equally separated into two classes (denoted as $\times$ and $\circ$, respectively).   $n$ points are chosen from each class as training samples (denoted as magenta $\times$ and $\circ$ ) and the rest as testing samples (denoted as blue $\times$ and $\circ$).  The principle component of $2n$ training points is computed by 2DPCA and R2DPCA.   In three random cases,  the computed principle components by two methods are plotted with the black lines, and the weighting vectors of R2DPCA are  $[0.7082,\ 0.2918]$, $[0.5407,\    0.4593]$  and $[0.5972,\   0.4028]$. 
 The variances (the larger the better) of the training set and the whole $4n$ points, under the projection of 2DPCA and R2DPCA,  are shown in Table \ref{t:variances}. 
  \begin{table}[!h]
  \caption{Variances in three random cases}\label{t:variances}
  \center
\begin{tabular}{c|cc|cc|cc}
  \hline
$4n$ &  \multicolumn{2}{c|}{  Variance of training points} 
      & \multicolumn{2}{c|}{  Variance of testing points} 
      &  \multicolumn{2}{c}{  Variance of the whole points}       \\ 
           &  {  2DPCA}     &  { R2DPCA}&  {  2DPCA}     &  { R2DPCA} &  {  2DPCA}     &  {R2DPCA}       \\ \hline
$20$  &      ${\bf 4.0757} $& $ 4.0161$    &      $ 2.9006 $& $ {\bf  2.9840}$        &      $3.4882$& ${\bf 3.5001}$            \\   
$500$  &      ${\bf 3.7994 } $& $    3.7956$    &      $ 3.6619$& $ {\bf   3.6741}$        &      $3.7306     $& ${\bf 3.7349}$            \\   
$1000$  &      ${\bf  3.8708} $& $  3.8626$    &      $ 4.4249 $& $ {\bf   4.4457}$        &      $4.1479 $& ${\bf 4.1541}$            \\   
\hline
\end{tabular}
\end{table}
 \begin{figure*}[!t]
  \begin{center}
 \includegraphics[height=0.4\textwidth,width=0.9\textwidth]{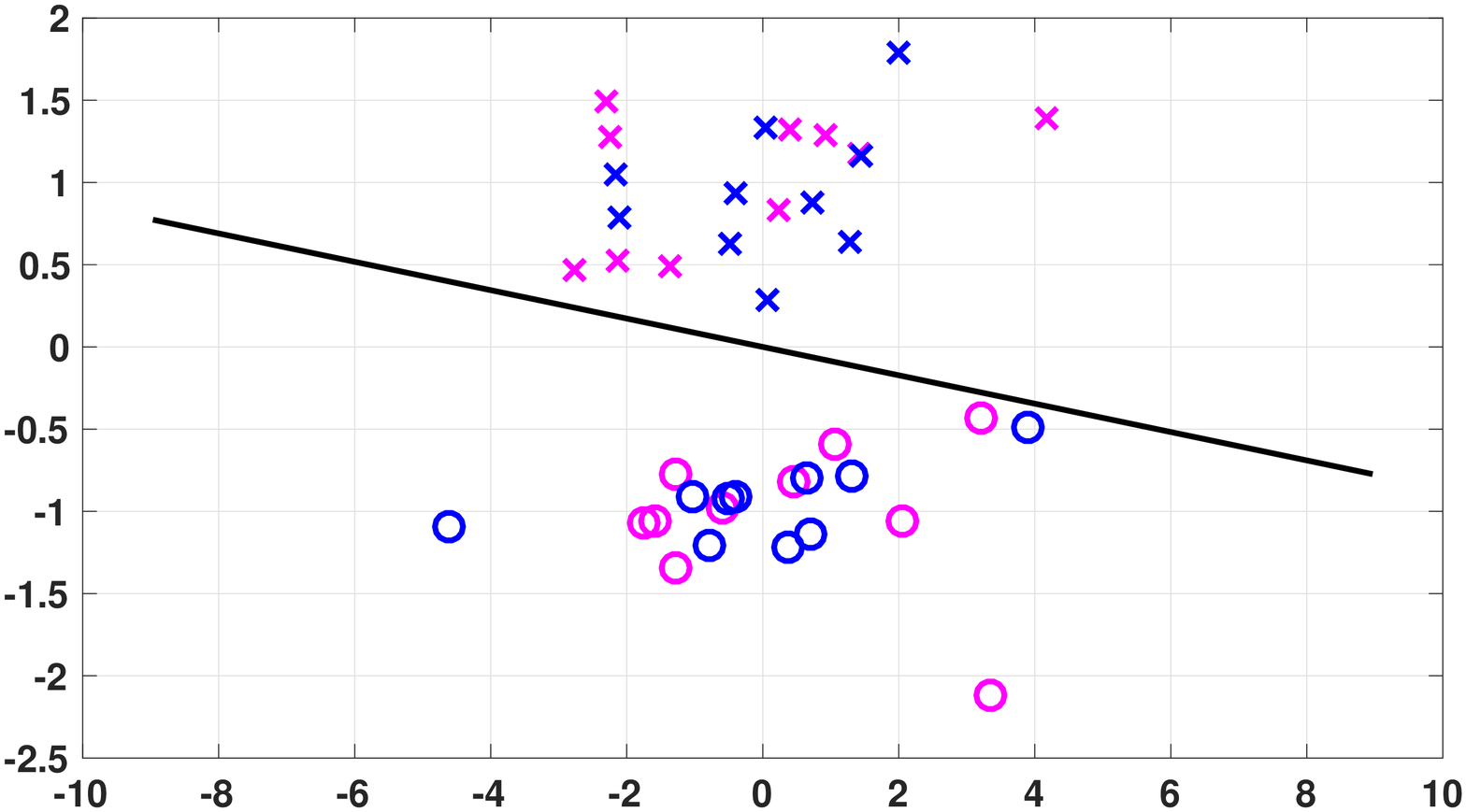}
 %
    \includegraphics[height=0.4\textwidth,width=0.9\textwidth]{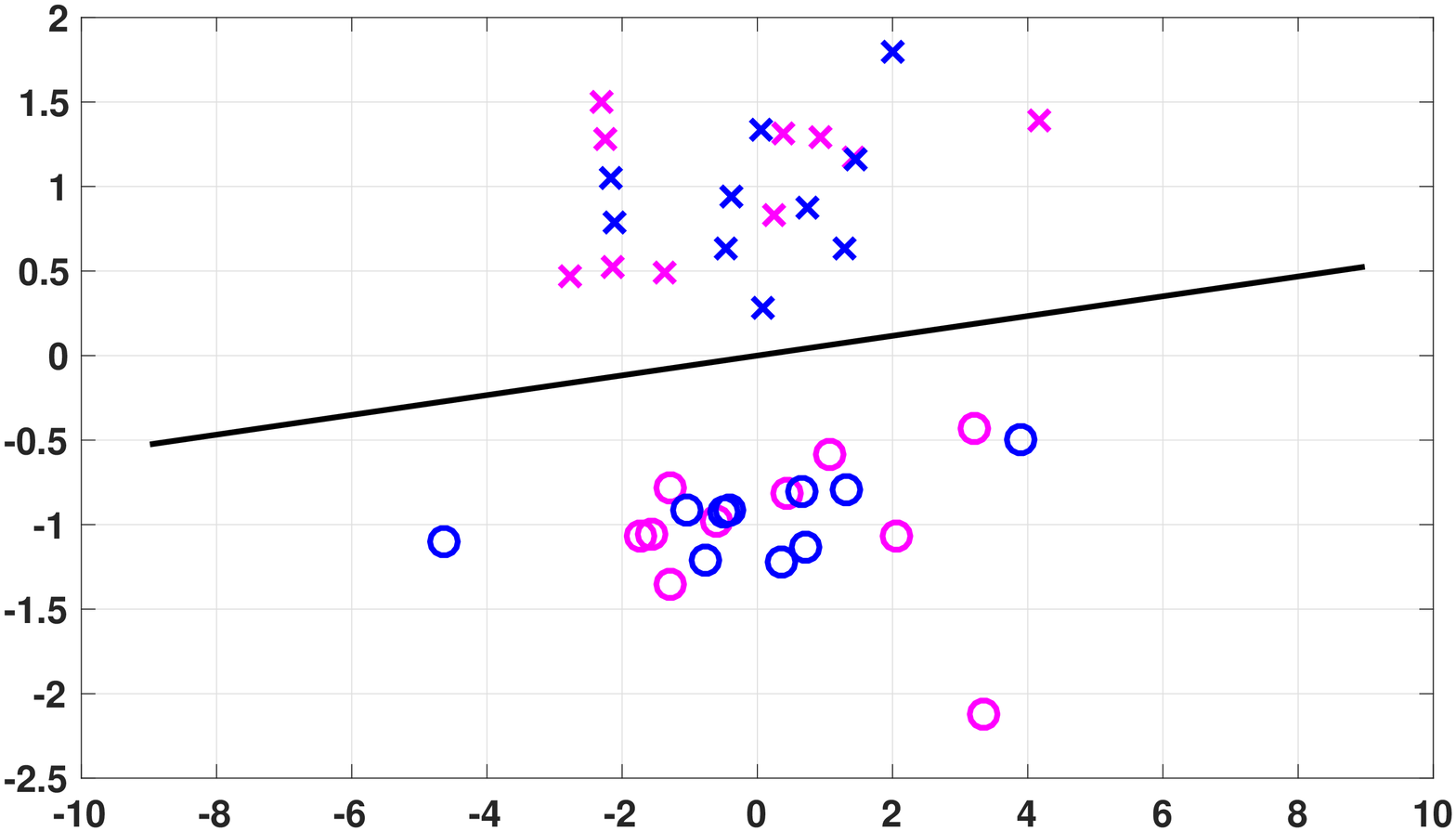}
\caption{
2DQPCA (Top) and R2DPCA(Bellow) in random case with $n=10$:  $4n$ randomly generated points  are equally separated into two classes (denoted as $\times$ and $\circ$, respectively). $n$ points from each class are chosen as training samples (denoted as magenta $\times$ and $\circ$ ) and the rest as testing samples (denoted as blue $\times$ and $\circ$).  The principle components of $2n$ training points are  plotted with the blue lines. 
%
 }
\end{center}
\label{f:toyexam}
\end{figure*}
\end{example}

\section{Conclusion}\label{s:conclusion}
This paper is a survey of recent development of 2DPCA. We present a general ridge regression model for  2DPCA  and variations by $L_p$ norm, with the improvement on  feature extraction   from both row and column directions.  To enhance the generalization ability, the relaxed 2DPCA  (R2DPCA) is proposed  with a general  ridge regression model.
The R2DPCA is a generalization of 2DPCA, 2DPCA-$L_1$ and G2DPCA, and has higher generalization ability. 
Since utilizing the label information, the R2DPCA can be seen as  a new supervised projection method, but it is totally different to  the two-dimensional linear discriminant analysis (2DLDA)\cite{ye05,lls08}.  
The  R2DPCA-based approaches for  face recognition and image reconstruction are also proposed and the selected  principle components are weighted to enhance the role of main components.
The properties and the effectiveness of proposed methods are verified  by  practical face image databases. 
In numerical experiments, R2DPCA has a better performance than 2DPCA, 2DPCA-$L_1$,G2DPCA, CNNs, DBNs, and DNNs.

\section*{Acknowledgments}
This paper is supported in part  by  National Natural Science
 Foundation of China  under grants 11771188 and 
 a Project Funded by the Priority Academic Program Development
  of Jiangsu Higher Education Institutions.




\section*{References}
\begin{thebibliography}{00}


\bibitem{yzfy04} J. Yang, D. Zhang, A. F. Frangi, J. Y. Yang (2004) \textsl{Two-dimensional PCA: A new approach to appearance-based face representation and recognition}, IEEE Trans. Pattern Anal. Mach. Intell., 26 (1), pp. 131-137.


\bibitem{zhzh05}Daoqiang Zhang and Zhi-Hua Zhou (2005), (2D)$^2$PCA: Two-directional two-dimensional PCA for efficient face representation and recognition, Neurocomputing, 69(1-3), pp. 224-231.

\bibitem{lpy10} X. Li, Y. Pang, and Y. Yuan (2010) \textsl{$L_1$-norm-based 2DPCA}, IEEE Trans. Syst., Man, Cybern. B, Cybern., 40 (4), pp. 1170-1175.

\bibitem{whwj13} H. Wang and J. Wang (2013) \textsl{2DPCA with $L_1$-norm for simultaneously robust and sparse modelling}, Neural Netw., 46, pp. 190-198.

\bibitem{wangj16} J. Wang (2016) \textsl{Generalized 2-D Principal Component Analysis by $L_p$-Norm for Image Analysis}, IEEE Trans. Cybern., 46 (3), pp. 792-803.

\bibitem{gxcdgl19}Q. Gao et al. (2019) \textsl{$R_1$-2-DPCA and face recognition},  IEEE Trans. Cybern., 49(4), pp. 1212-1223.

\bibitem{yuwu17} D. Yu, X.-J. Wu (2017) 2DPCANet: a deep leaning network for face recognition, Multimed. Tools Appl., 4, pp. 1-16.

\bibitem{lwk19}Y.-K. Li, X.-J. Wu, and J. Kittler (2019) L1-2D2PCANet: a deep learning network for face recognition,  J. of Electronic Imaging, 28(2), pp. 023016 (20 March 2019). https://doi.org/10.1117/1.JEI.28.2.023016


\bibitem{Jolliffe04} I. Jolliffe (2004) \textsl{Principal Component Analysis}, New York, NY, USA: Springer.

\bibitem{TP91} M. Turk, A. Pentland (1991) \textsl{Eigenfaces for recognition}, J. Cogn. Neurosci., 3 (1), pp. 71-86.

\bibitem{siki87} L. Sirovich, M. Kirby (1987) \textsl{Low-dimensional procedure for characterization of human faces}, J. Optical Soc. Am. 4, pp. 519-524.

\bibitem{kisi90} M. Kirby,  L. Sirovich (1990)
\textsl{Application of the karhunenloeve procedure for the characterization of human faces},
IEEE Trans. Pattern Anal. Mach. Intell., 12 (1), pp. 103-108.

\bibitem{tupe91} M. Turk, A. Pentland (1991) \textsl{Eigenfaces for recognition. J. Cognitive Neurosci}, 3(1),  pp. 71-76.

\bibitem{zhya99} L. Zhao, Y. Yang (1999) \textsl{Theoretical analysis of illumination in PCA-based vision systems},  Pattern Recogn.,  32(4), pp. 547-564.

\bibitem{pent00}  A. Pentland (2000)
\textsl{Looking at people: sensing for ubiquitous and wearable computing},
IEEE Trans. Pattern Anal. Mach. Intell., 22 (1), pp. 107-119.

\bibitem{keka05} Q. Ke and T. Kanade (2005) \textsl{Robust $L_1$ norm factorization in the presence of outliters and missing data by alternative convex programming}, Proc. IEEE Conf. Comput. Vis. Pattern Recognit., 1, San Diego, CA, USA, pp. 739-746.

\bibitem{dzhz06} C. Ding, D. Zhou, X. He, and H. Zha (2006) \textsl{$R_1$-PCA: Rotational invariant $L_1$-norm principal component analysis for robust subspace factorization}, Proc. 23rd Int. Conf. Mach. Learn., Pittsburgh, PA, USA, pp. 281-288.

\bibitem{kwak08} N. Kwak (2008) \textsl{Principal component analysis based on $L_1$-norm maximization}, IEEE Trans. Pattern Anal. Mach. Intell., 30 (9), pp. 1672-1680.

\bibitem{zht06} H. Zou, T. Hastie, and R. Tibshirani (2006) \textsl{Sparse principal component analysis}, J. Comput. Graph. Stat., 15 (2), pp. 265-286.

\bibitem{agjl07} A. d'Aspremont, L. EI Ghaoui, M. I. Jordan, and G. R. Lanckriet (2007) \textsl{A direct formulation for sparse PCA using semidefinite programming}, SIAM Rev., 49 (3), pp. 434-448.

\bibitem{shhu08} H. Shen and J. Z. Huang (2008) \textsl{Sparse principal component analysis via regularized low rank matrix approximation}, J. Multivar. Anal., 99 (6), pp. 1015-1034.

\bibitem{wth09} D. M. Witten, R. Tibshirani, and T. Hastie (2009) \textsl{A penalized matrix decomposition, with applications to sparse principal components and canonical correlation analysis}, Biostatistics, 10 (3), pp. 515-534.

\bibitem{mzx12} D. Meng, Q. Zhao, and Z. Xu (2012) \textsl{Improve robustness of sparse PCA by $L_1$-norm maximization}, Pattern Recognit., 45 (1), pp. 487-497.

\bibitem{kwak14} N. Kwak (2014) \textsl{Principal component analysis by $L_p$-norm maximization}, IEEE Trans. Cybern., 44 (5), pp. 594-609.

\bibitem{lxzzl13} Z. Liang, S. Xia, Y. Zhou, L. Zhang, and Y. Li (2013) \textsl{Feature extraction based on $L_p$-norm generalized principal component analysis}, Pattern Recognit. Lett., 34 (9), pp. 1037-1045.





\bibitem{jlz17} Z. Jia, S. Ling,  M. Zhao (2017) Color  two-dimensional principal component analysis  for face recognition based on quaternion model,   LNCS,  vol. 10361,   pp. 177-189. 
\bibitem{zjg17} M. Zhao, Z. Jia, D. Gong (2018) \textsl{Sample-relaxed two-dimensional color principal component analysis for face recognition and image reconstruction},  arXiv.org/cs /arXiv:1803.03837v1,  10 Mar 2018.

\bibitem{jwl13}  Z. Jia,   M. Wei,   S. Ling (2013)  A new structure-preserving method
 for quaternion Hermitian eigenvalue problems,  J.  Comput. Appl. Math. 239, pp. 12-24.
\bibitem{mjb18}R. Ma, Z. Jia, Z. Bai (2018) A structure-preserving Jacobi algorithm for quaternion Hermitian eigenvalue problems, Comput.  Math. Appl., 75(3),  pp. 809-820.
\bibitem{jmz17}Z. Jia, R. Ma, M. Zhao   (2017) 
 A New Structure-Preserving Method for Recognition of Color Face Images, Computer Science and Artificial Intelligence , pp. 427-432.
  \bibitem{jwzc18} Z. Jia,   M. Wei,  M. Zhao, Y. Chen (2018) A new real structure-preserving quaternion QR algorithm, J.  Comput. Appl. Math.  343, pp. 26-48.
 \bibitem{jns18b} Z. Jia, M.K. Ng, G. Song (2018) Lanczos method for large-scale quaternion singular value decomposition, Numer. Algorithms, 08 November 2018.  https://doi.org/10.1007/s11075-018-0621-0
 \bibitem{jns18a} Z. Jia, M.K. Ng,  and G. Song (2019)  Robust Quaternion Matrix Completion with Applications to Image Inpainting,  Numer. Linear Algebra  Appl.,  DOI:10.1002/nla.2245. http://www.math.hkbu.edu.hk/~mng/quaternion.html
 \bibitem{Mackey08} L. Mackey (2008) \textsl{Deflation methods for sparse PCA}, Proc. Adv. Neural Inf. Process. Syst., 21, Whistler, BC, Canada., pp. 1017-1024.
 
 \bibitem{bhk97} P.N. Belhumeur,  J. Hespanda,  D. Kriegeman (1997)  Eigenfaces vs Fisherfaces:
Recognition using class specific linear projection. IEEE Trans. Pattern Anal.
Machine Intell. 19(7), pp. 711-720.

\bibitem{ye05} J. Ye (2005) Characterization of a family of algorithms for generalized discriminant
analysis on undersampled problems, Machine Learning Res., 6, pp. 1532-4435.

\bibitem{hopa04} P. Howland,  H. Park (2004)   Generalized discriminant analysis using the generalized
singular value decomposition, IEEE Trans. Pattern Anal. Machine Intell. 8, pp. 995-1006.
 


\bibitem{yeli05} J. Ye,  Q. Li (2005)  A two-stage discriminant analysis via QR decomposition, IEEE
Trans. Pattern Anal. Machine Intell., 27(6), pp. 929-941.

 \bibitem{lls08} Z.Z. Liang,  Y.F. Li,   P.F. Shi (2008) A note on two-dimensional linear discriminant analysis, Pattern Recognit. Lett. 29,  pp. 2122-2128.
 
 
\bibitem{kora05}  S. Kongsontana, Y. Rangsanseri (2005)  Face recognition using 2DLDA algorithm, In:
Proc. 8th Internat. Symp. Signal Process. Appl.,  pp. 675-678.

\bibitem{xsa05} H. Xiong,  M.N.S. Swamy, M.O.  Ahmad  (2005) Two-dimensional FLD for face
recognition. Pattern Recognition 38 (7),  1121-1124.

\bibitem{yyfz03}J. Yang,  J.Y. Yang,  A.F. Frangi,  D.  Zhang (2003)  Uncorrelated projection
discriminant analysis and its application to face image feature extraction,
Internat. J. Pattern Recognition Artificial Intell. 17 (8), pp. 1325-1347.

\bibitem{liyu05}  M. Li, B. Yuan (2005)  2D-LDA: A novel statistical linear discriminant analysis for
image matrix. Pattern Recognition Lett. 26 (55), pp. 527-532.

\bibitem{cckkl06} D. Cho, U. Chang,  K. Kim,  B. Kim,  S. Lee  (2006)   (2D)2DLDA for efficient face
recognition. LNCS 4319, pp. 314-321.
 
 
 

 

\bibitem{palm12} R.B.Palm. Prediction as a candidate for learning deep hierarchical models of data.Technical University of Denmark, 2012.
\bibitem{nielsen13}  M. Nielsen. Neural Networks and Deep Learning[online]. 2013. http://neuralnetworksanddeeplearning.com


\end{thebibliography}


\end{document}